\documentclass[twoside]{article}

\usepackage[preprint]{aistats2022}

\usepackage[round]{natbib}

\usepackage[utf8]{inputenc} % allow utf-8 input
\usepackage[T1]{fontenc}    % use 8-bit T1 fonts
\usepackage[hyperfootnotes=false]{hyperref}       % hyperlinks
\usepackage{url}            % simple URL typesetting
\usepackage{booktabs}       % professional-quality tables
\usepackage{amsfonts}       % blackboard math symbols
\usepackage{nicefrac}       % compact symbols for 1/2, etc.
\usepackage{microtype}      % microtypography
\usepackage{xcolor}         % colors
\usepackage{subcaption}     % subfigs
\usepackage{enumitem}
\usepackage{stfloats}      % positioning floats with two cols
\usepackage{placeins}      % for \FloatBarrier
\usepackage{mathtools,amsmath,amssymb}

\usepackage[export]{adjustbox}  % for figs in appendix E

\newcommand{\x}{\mathbf{x}}
\newcommand{\w}{\mathbf{w}}
\newcommand{\X}{\mathbf{X}}
\newcommand{\y}{\mathbf{y}}

\newcommand{\p}{\mathbf{p}}
\newcommand{\f}{\mathbf{f}}

\newcommand{\lrbracket}[3]{\left#1 #3 \right#2}
\newcommand{\lmrbracket}[5]{\left#1 #4 \middle#2 #5 \right#3}

\renewcommand{\b}{\lrbracket{(}{)}}
\newcommand{\bc}{\lmrbracket{(}{\vert}{)}}
\newcommand{\sqb}{\lrbracket{[}{]}}

\renewcommand{\L}{\mathcal{L}}

\newcommand{\Lnoaug}{\mathcal{L}_\text{noaug}}
\newcommand{\Ladd}{\mathcal{L}_\text{add}}
\newcommand{\Llogprob}{\mathcal{L}_\text{loss}}
\newcommand{\Llogits}{\mathcal{L}_{\text{logits}}}
\newcommand{\Lprob}{\mathcal{L}_{\text{prob}}}

\newcommand{\Lhlogprob}{\hat{\mathcal{L}}_\text{loss}}
\newcommand{\Lhlogits}{\hat{\mathcal{L}}_{\text{logits},K}}
\newcommand{\Lhprob}{\hat{\mathcal{L}}_{\text{prob},K}}

\newcommand{\LhlogitsKone}{\hat{\mathcal{L}}_{\text{logits};1}}
\newcommand{\LhprobKone}{\hat{\mathcal{L}}_{\text{prob};1}}

\newcommand{\Ktrain}{K_\text{train}}
\newcommand{\Ktest}{K_\text{test}}

\let\P\relax
\DeclareMathOperator{\P}{P}
\DeclareMathOperator{\Q}{Q}
\DeclareMathOperator{\E}{\mathbb{E}}
\newcommand{\Plogits}{\P_\text{logits}}
\newcommand{\Pprob}{\P_\text{prob}}
\newcommand{\Pnoaug}{\P_\text{noaug}}

\DeclareMathOperator{\const}{const}
\DeclareMathOperator{\softmax}{softmax}

\newcommand{\finv}{\f_\text{inv}}
\newcommand{\pinv}{\p_\text{inv}}

\newcommand\blfootnote[1]{%
  \begingroup
  \renewcommand\thefootnote{}\footnote{#1}%
  \addtocounter{footnote}{-1}%
  \endgroup
}

\newcommand{\aistatstitlenovskip}[1]{
 \hsize\textwidth
  \linewidth\hsize \toptitlebar {\centering
  {\Large\bfseries #1 \par}}
 \bottomtitlebar
}
\newcommand{\tsum}{{\textstyle \sum}}

\newcommand{\myeqref}[1]{(Eq.~\ref{#1})}

\author{
  Seth Nabarro* \\
  Department of Computing \\
  Imperial College London \\
  London, SW7 2BX, UK \\
  \texttt{seth.nabarro09@imperial.ac.uk}
  \And
  Stoil Ganev*\\
  Department of Computer Science\\
  University of Bristol \\
  Bristol, BS8 1UB, UK \\
  \And
  Adrià Garriga-Alonso\\
  Department of Engineering\\
  University of Cambridge \\
  Cambridge, CB2 1PZ, UK \\
  \texttt{ag919@cam.ac.uk}
  \And
  Vincent Fortuin\\
  Department of Computer Science,\\ 
  ETH Zürich,\\ 
  Zürich,  Switzerland\\
  \And
  Mark van der Wilk\textdagger \\
  Department of Computing \\
  Imperial College London \\
  London, SW7 2BX, UK \\
  \texttt{m.vdwilk@imperial.ac.uk}
  \And
  Laurence Aitchison\textdagger\\
  Department of Computer Science\\
  University of Bristol \\
  Bristol, BS8 1UB, UK \\
  \texttt{laurence.aitchison@gmail.com} \\
}

\begin{document}
\runningtitle{Data augmentation in Bayesian neural networks and the cold posterior effect}
\runningauthor{Nabarro, Ganev, Garriga-Alonso, Fortuin, van der Wilk, Aitchison}
\twocolumn[
\aistatstitle{Data augmentation in Bayesian neural networks\\and the cold posterior effect}

\aistatsauthor{Seth Nabarro* \And Stoil Ganev* \And  Adrià Garriga-Alonso}

\aistatsaddress{Department of Computing \\
  Imperial College London \\
  London, SW7 2BX, UK \\
  \texttt{seth.nabarro09@imperial.ac.uk} \And  
  Department of Computer Science\\
  University of Bristol \\
  Bristol, BS8 1UB, UK \\ \And 
  Department of Engineering\\
  University of Cambridge \\
  Cambridge, CB2 1PZ, UK \\
  \texttt{ag919@cam.ac.uk}}
  
\aistatsauthor{Vincent Fortuin \And Mark van der Wilk\textdagger \And Laurence Aitchison\textdagger}
\aistatsaddress{
  Department of Computer Science,\\ 
  ETH Zürich,\\ 
  Zürich,  Switzerland\\ \And
  Department of Computing \\
  Imperial College London \\
  London, SW7 2BX, UK \\
  \texttt{m.vdwilk@imperial.ac.uk}\\
  \And 
  Department of Computer Science\\
  University of Bristol \\
  Bristol, BS8 1UB, UK \\
  \texttt{laurence.aitchison@gmail.com}
  }

]

\begin{abstract}
Bayesian neural networks that incorporate data augmentation implicitly use a ``randomly perturbed log-likelihood [which] does not have a clean interpretation as a valid likelihood function'' (Izmailov et al. 2021).
Here, we provide several approaches to developing principled Bayesian neural networks incorporating data augmentation. 
We introduce a ``finite orbit'' setting which allows likelihoods to be computed exactly, and give tight multi-sample bounds in the more usual ``full orbit'' setting.
These models cast light on the origin of the cold posterior effect.
In particular, we find that the cold posterior effect persists even in these principled models incorporating data augmentation.
This suggests that the cold posterior effect cannot be dismissed as an artifact of data augmentation using incorrect likelihoods.
%Data augmentation has recently been suggested as the origin of the cold posterior effect in Bayesian neural networks (Izmailov et al. 2021).
%However, this argument is problematic as long as we do not have a principled approach to integrating data augmentation within Bayesian neural networks.
%We give several such approaches, 
%The cold posterior effect persists even in these principled models, suggesting that the cold posterior effect cannot be dismissed as an artifact of data curation.
%Indeed, 
%Data augmentation is a highly effective approach for improving performance in deep neural networks. The standard view is that it creates an enlarged dataset by adding synthetic data, which raises a problem when combining it with Bayesian inference: how much data are we really conditioning on? This question is particularly relevant to recent observations linking data augmentation to the cold posterior effect. We investigate various principled ways of finding a log-likelihood for augmented datasets. Our approach prescribes augmenting the same underlying image multiple times, both at test and train-time, and averaging either the logits or the predictive probabilities. Empirically, we observe the best performance with averaging probabilities. While there are interactions with the cold posterior effect, neither averaging logits or averaging probabilities eliminates it.

\blfootnote{* equal contribution}\blfootnote{\textdagger{} equal contribution}

\end{abstract}

\section{INTRODUCTION}
\label{sec:intro}
%It is well known that Bayesian inference in combination with Bayes decision theory should give optimal performance in a well-specified model .
The cold posterior effect \citep[CPE;][]{wenzel2020good} is the surprising  observation that performance in neural networks is not optimal when we use the usual Bayesian posterior \citep{kolmogorov1950foundations,savage1954foundations,jaynes2003probability},
\begin{align}
  \P\bc{\w}{\y, \X} \propto \P\b{\w} \P\bc{\y}{\w, \X}
\end{align}
where $\w$ are the neural network weights, $\X$ is all inputs (typically images), and $\y$ is all outputs (typically class labels).
Instead, we get better performance when using a ``cold'' posterior, i.e. the posterior taken to the power of $1/T$ where $T<1$,
\begin{align}
  \Q\b{\w} &\propto \b{\P\b{\w} \P\bc{\y}{\w, \x}}^{1/T}.
\end{align}
The origin of the CPE is by now highly contentious, with three leading potential explanations \citep{noci2021disentangling}.
The first hypothesis is that the process of data curation for popular datasets such as CIFAR-10 and ImageNet \citep{krizhevsky2009learning,deng2009imagenet} involves multiple annotators agreeing upon the label for each image.
In that case, there are in effect multiple labels for each image, which inflates the likelihood (but not the prior) term in the cold posterior \citep{adlam2020cold,aitchison2020cold}.
Second, the prior is always misspecified, and prior misspecification is known to induce cold posterior-like effects in specific (non-neural network) models \citep{grunwald2012safe,grunwald2017inconsistency}, which might give an explanation for the CPE in neural networks \citep{wenzel2020good,fortuin2021bayesian}. % This is supported by \citet{noci2021disentangling}, who show that the CPE can be more pronounced when less data is available, and therefore the prior is more influential.
%while a variety of improved priors have been tried and they do improve performance, these improvements are typically small in comparison to the magnitude of the cold-posterior effect for large-scale network of interest. 
However, \citet{fortuin2021bayesian} showed that better priors do not always reduce the size of the CPE, but can actually increase it.  
In particular, they found that incorporating spatial correlations in convolutional filters improved the performance of a ResNet trained on CIFAR-10, but also increased the magnitude of the CPE. %, which shows that better priors do not necessarily reduce the CPE. % see e.g.\ \citet{ober2020global} and \citet{fort2021drawing} (specifically, the ResNet for CIFAR-10 in that paper).
%As such, \citet{aitchison2020cold} argues that there is as of yet no evidence that a prior could be found that closes the cold-posterior effect.
Third, there is the possibility that the CPE is an artifact of data augmentation \citep[DA; ][]{wenzel2020good,izmailov2021bayesian}, as DA gives a ``randomly perturbed log-likelihood [which] does not have a clean interpretation as a valid likelihood function'' \citep{izmailov2021bayesian}. This is supported by observations in which the CPE only exists with DA, and disappears without DA \citep{wenzel2020good,fortuin2021bayesian,izmailov2021bayesian}.
Of course it is quite possible that practical CPEs arise from a complex combination of these causes \citep{aitchison2020cold,noci2021disentangling}.

In spite of this controversy, recent work on the CPE agrees that it is important to investigate integrating DA with Bayesian neural networks (BNNs), and to examine the interaction with the CPE.
From \citet{noci2021disentangling}: ``It remains an interesting open problem how to properly account for data augmentation in a Bayesian sense.''
And from \citet{izmailov2021bayesian}: ``Data augmentation cannot be naively incorporated in the Bayesian neural network model.'' And ``We leave incorporating data augmentation ... as an exciting direction of future work.''

Perhaps the most common understanding of the interaction between the CPE and DA in BNNs is that DA increases the effective dataset size.
From \citet{izmailov2021bayesian}: ``intuitively, data augmentation increases the amount of data observed by the model, and should lead to higher posterior contraction''.  
From \citet{osawa2019practical}: ``DA increases the effective sample size''.
%In footnote 3 in \citet{wenzel2020good}, they note ``For \citep{heek2019bayesian} we communicated with the authors, and tempering arises due to overcounting data by a factor of 5, approximately justified by data augmentation, corresponding to T = 1/5.''
From, \citet{noci2021disentangling}: ``while data augmentation may increase the amount of data seen by the model, that increase is certainly not equal to the number of times each data point is augmented (after all, augmented data is not independent from the original data).''

In this paper, we first give a formal argument that the notion that DA increases the effective dataset size is flawed.
Second, we consider issues with the current training objectives for Bayesian generative models which incorporate DA \citep{van2018learning,wenzel2020good}. %, and which have been investigated empirically in the non-Bayesian setting \citep[e.g.][]{krizhevsky2012imagenet,lyle2020benefits,fort2021drawing}.
In particular, \citet{wenzel2020good} proposed a single-sample training objective that \citet{izmailov2021bayesian} regarded as a ``randomly perturbed log-likelihood [which] does not have a clean interpretation as a valid likelihood function'', and \Citet{van2018learning} proposed a training objective that only works for quadratic log-likelihoods (i.e.\ Gaussians or P\'{o}lya-Gamma approximations).
%the invariance construction of \Citet{van2018learning} for BNNs. This prescribes a generative model for the labels in which we take a single image, augment it multiple times, apply the neural network to each augmentation, then form an overall prediction by averaging the outputs.
%Similar algorithms have been investigated empirically in a non-Bayesian setting \citep[e.g.][]{krizhevsky2012imagenet,lyle2020benefits,fort2021drawing}.
We show that multi-sample estimators which average network logits or probabilities form lower bounds on the intractable log-likelihoods of principled Bayesian models. These bounds are tighter than the single-sample estimators used previously in the Bayesian setting \citep[e.g.][]{wenzel2020good} and can be applied to a broad class of likelihood functions \Citep[unlike][]{van2018learning}.
Third, we introduce a ``finite orbit'' setting with a small number of admissible augmentations which allows us to compute \textit{exact} log-likelihoods. Fourth, we give the natural generalizations beyond classification to arbitrary outputs (Appendix~\ref{sec:app:generalization}). %, and compare their justifications and behaviours (Appendix~\ref{sec:app:perspectives}).
Fifth, we show that the CPE persists even when using these principled DA likelihoods (it certainly is the case that the CPE could have been an artifact of loose bounds arising from previous single-sample estimators). 
Finally, we consider the consequences for explanations of the CPE. 
In particular, we can no longer conclude that the CPE is an artifact resulting from DA giving an invalid likelihood function \citep{izmailov2021bayesian}, as we have a principled Bayesian model incorporating DA. 
In the finite orbit setting, we can compute the exact likelihood, while in the usual setting there is in principle a deterministic likelihood with a clean interpretation as a valid likelihood function, but as this is difficult to evaluate in practice we use a tight multi-sample lower bound.
%In principle, this model has a 

Note that in the remainder of the paper, we will follow \citet{izmailov2021bayesian} in regarding models with loose, single-sample bounds as ``unprincipled'' (from \citet{izmailov2021bayesian}, the ``randomly perturbed log-likelihood does not have a clean interpretation as a valid likelihood function'').
In contrast, we term models using our exact log-likelihoods or our tight multi-sample bounds as being ``principled''.

\section{BACKGROUND\label{sec:background}}
\subsection{Data augmentation}
\label{sec:background:da}

In supervised learning, we are interested in learning some unknown functional relationship from example input-output pairs $(\x_i, y_i), i=1,\ldots,N$. 
%We can usually expect better performance when more examples are given \citep{loog2019erm}.
Usually, we have information about some form of invariance, i.e.\ the knowledge that the function does not change its output for certain transformations of the input.
These might occasionally be true invariances, such as the identity of a molecule being invariant to rotations.
But in most settings, these are so-called ``soft'' invariances or ``insensitivities'' \citep{van2018learning}. 
For instance the class label for an image should not change due to small translations/crops of that image (but might change if we radically alter the image).
The most basic form of data augmentation takes advantage of this information by transforming, or augmenting, the inputs and copying the output value, to create additional input-output pairs which are then included in training.
Often, the amount of additional ``augmented data'' can be unbounded, for example when allowable transformations are specified in a continuous range, e.g.\ rotations.
This simple procedure has been very successful in improving performance in a wide variety of machine learning methods \citep{loosli2007training,krizhevsky2012imagenet,bishop2006pattern}, and recent work has analysed the effect of data augmentation on invariances in the learned functions \citep{dao2019kernelda,chen2020groupda,lyle2020benefits}.

\subsection{Bayesian inference}
\label{sec:background:bayes_inf}

Bayesian inference allows us to infer a distribution over neural network weights, which incorporates uncertainty induced by having finite data.
Bayes prescribes a strict procedure for updating beliefs about unknown quantities in light of observed data.
The model is specified by a prior on the weights $\P(\w)$ and a log-likelihood, $\sum_{i=1}^N \L^i(y_i;\w)$. 
Thus, the log-posterior is given by
\begin{align}
\log\P\bc{\w}{\X, \y} &= \log \P\b{\w} + \sum_{i=1}^N \L^i(\w) + \const.
\end{align}
Without augmentation, the multi-class classification log-likelihood is given by
\begin{align} 
  \Lnoaug^i(y_i;\w)&=\log \Pnoaug\bc{y_i}{\w, \x_i} \nonumber\\
  &= \log \softmax_{y_i} \f(\x_i; \w) ,  \label{eq:def:Lnoaug}
\end{align}
where $\f(\x_i; \w)$ is the neural network outputs, which are treated as the logits for a categorical distribution.

\section{METHODS}
\label{sec:meth}

\subsection{Arguments against previous approaches to DA and the CPE}

One idea noted in Sec.~\ref{sec:intro} is that each augmented image can be understood as its own datapoint, in which case DA increases the effective dataset size.
Taking $A$ augmented images, $\x'_{i; a}$, we can write the resulting log-likelihood for a single underlying image as,
\begin{align}
  \Ladd^i(y_i;\w)=\sum_{a=1}^A\log \softmax_{y_i} \f(\x'_{i;a}; \w).
\end{align}
%Here $A$ is the total number of augmentations, and $a$ indexes different augmentations of the same underlying image.
The first issue with this approach is that the ``correct'' number of augmentations can be very difficult to define.
Indeed, in the case of e.g.\ rotations, there are an uncountably infinite number of possible augmentations even within a narrow range, perhaps implying that we should take $A=\infty$, which is clearly pathological as it results in ignoring the prior.
%However, when doing inference we are free to choose the prior and likelihood, but not the way they are combined to find the posterior.
%Changing the dataset size through augmentation is inconsistent with the Bayesian principle of requiring one likelihood for each ``true'' datapoint.
%%While we might try to interpret $\Ladd^i(y_i;\w)$, but this does not make sense, because the log-likelihood has to form a valid distribution over output labels, and this ``distribution'' does not in general normalise,
%%In this context, $\Ladd^i(y_i;\w)$ is not a valid likelihood for the underlying ``true'' datapoint, because it does not in general form a distribution that normalises,
%Suppose we were to go ahead anyway, treat each augmented image as separate datapoints and thus use $\Ladd$,
%
%\begin{align}
%  \Ladd^i(y_i;\w)=\sum_{a=1}^A\log \softmax_{y_i} f(\x'_{i;k}; \w)
%\end{align}
%
%By doing this, we are assuming that all $(\x'_{i,a},y_i)$ and $(\x_i,y_i)$ are iid, when in fact we have generated $\x'_{i,a}$ directly from $\x_{i}$. 
%This is likely to result in a posterior which is overconfident, as it assumes $A$ times more data points than we really have.
%The infinite augmentation case is particularly problematic, as it results in ignoring the prior, which is clearly unjustified given all our examples are generated from a finite set of underlying observations.
%
This issue is really hinting at a deeper problem: in treating each augmented image as a separate datapoint, we have implicitly assumed that the labels for each augmented image are independent.
This could be achieved if we took each augmentation of the same underlying image and presented them to a different human annotator.
However, that is not what happens in practice.
Usually it is only the underlying unaugmented image, $\x_i$, that is labelled by a human, and that single label is assumed to apply to all augmented images, $\x'_{i; a}$.
As such, the labels for different augmentations of the same underlying input are not independent, and an approach (such as this one) which assumes they are cannot be valid.

Perhaps the most standard practical approach to using DA in BNNs is to take a pre-existing algorithm, and replace the underlying non-augmented image, $\x_i$, with a randomly augmented image, $\x_i'$.
However, if we consider $\log \softmax_{y_i} \f(\x'_i; \w)$ to be the log-likelihood, then randomness in the augmented image, $\x'_i$ gives a ``randomly perturbed log-likelihood [which] does not have a clean interpretation as a valid likelihood function'' \citep{izmailov2021bayesian}.
Perhaps a more fundamental issue is that ultimately, the resulting algorithms target the averaged (negative) loss (Appendix~\ref{sec:app:vi_sgld}),
\begin{align}
  \Llogprob^i\b{y_i; \w} &=  \E\sqb{\log \softmax_{y_i} \f(\x'_i; \w)} .
\end{align}
This approach is convenient, as a single sample from the augmentation distribution can provide an unbiased estimate $\Lhlogprob=\log \softmax_{y_i} \f(\x'_i; \w)$, so it is used explicitly in some settings \citep[e.g.][]{benton2020learning}.
Importantly though, a valid likelihood should arise from a valid distribution over labels, and should therefore normalize if we sum over labels.
For instance, without augmentation,
\begin{align}
  \label{eq:likelihood_normalize}
  1 &= {\textstyle \sum}_{y_i=1}^Y \exp \Lnoaug^i(y_i;\w).
\end{align}
In practice, a likelihood which normalizes to a constant other than one is sufficient when doing inference with e.g. MCMC. However, the normalization constant for $\Llogprob(\cdot;\w)$ may vary with input location: ${\tsum}_{y_i=1}^Y\exp\Llogprob^i(y_i;\w)=Z(\x_i)$, and thus $\Llogprob^i(y_i;\w)$ is not a valid log-likelihood.

\subsection{Tight lower bounds on the log-likelihood of principled DA models}
\label{sec:tighter}
To obtain a principled log-likelihood incorporating DA, we cannot take the standard approach of just using augmented data in a pre-existing algorithm.
Instead, we need to build DA into the probabilistic generative model for labels.
To do this, we take inspiration from previous work which builds functional invariance into Gaussian process (GP) priors \citep{van2018learning,kondor2008group,ginsbourger2012argumentwise,ginsbourger2013invariances}. The approach constructs an (exactly) invariant function $h(\cdot;\w)$ by averaging a non-invariant function $g(\cdot;\w)$ over the distribution of interest, in this case the distribution over augmentations given input $\P\bc{\x'}{\x}$
\begin{align}
    h(\x_i;\w)=\int g(\x_i';\w)\P\bc{\x_i'}{\x_i}d\x_i'.
    \label{eq:invariance}
\end{align}
We thus consider models in which the output class label is given by averaging the (non-invariant) neural network over a distribution of all possible augmentations for each input. %, while ensuring that normalization (Eq.~\ref{eq:likelihood_normalize}) is satisfied. 
This simultaneously ensures that the likelihood normalizes \myeqref{eq:likelihood_normalize}, the averaged output is invariant to the augmentation transformations and circumvents the need to find the ``true'' number of augmentations.
Note that \myeqref{eq:invariance} implies the algorithm used in much non-Bayesian work, which averages the neural network output over multiple augmentation of each input \citep[e.g.][]{krizhevsky2012imagenet,he2015delving,szegedy2015going,simonyan2014very,foster2020improving}.

%In doing so, we circumvent the need to find the ``true'' number of examples in our augmented dataset.
% Importantly, averaging over an augmentation group (such as horizontal flips) creates an invariance, so our averaging methods are deeply related to a view of data-augmentation as introducing function-space invariance, which follows the method developed for GPs by \citet{van2018learning}.

%More specifically, the method constructs a function invariant to augmentations $\finv(\cdot;\w)$, by integrating a non-invariant function $f(\cdot;\w)$ over $\P(\x'|\x)$
%\begin{equation}
%    \finv(\x;\w)=\int_{\mathcal{X}'} f(\x';\w)\P(\x'|\x)d\x'
%    \label{eq:invariant_construct}
%\end{equation}
%This allows us to do inference on the weights of $\finv(\cdot;\w)$ with a minor amendment to existing inference algorithms: average the values of $f(\cdot;\w)$ over multiple augmentation samples for each input. See Section~\ref{sec:meth:invariance} for details.
%
%We note that the approach developed by \cite{van2018learning} is arguably the first formulation of truly Bayesian data augmentation, in that case for GP inference. The work also describes a method of learning function-space invariances using the marginal likelihood, however this is beyond the scope of this paper. 

In a classification setting, we have a choice as to which quantity we average: logits (equal to the neural network outputs) or predictive probabilities,
%, and hence at which level we enforce the invariance: at the level of logits or predictive probabilities, that is,
\begin{align}
  \pinv(\x_i;\w)&= \E\sqb{\softmax \f(\x_i'; \w)},\\
  \finv(\x_i;\w)&= \E\sqb{\f(\x_i'; \w)}.
\end{align}
where we take expectations over $\P(\x_i'|\x_i)$.
Remember that $\f(\x_i'; \w)$ is the (vector-valued) neural network output for an augmented input, which is used as the logits in classification, so $\finv(\x_i; \w)$ is the outputs averaged over all augmentations of the same underlying image. 
Likewise, $\pinv(\x_i; \w)$ is the vector of probabilities given by averaging the predicted probabilities over augmentations.
These quantities, $\finv(\x_i; \w)$ and $\pinv(\x_i; \w)$ are subscripted ``inv'' to denote that averaging over augmentations can give invariances in $\finv(\x_i; \w)$ and $\pinv(\x_i; \w)$ that are not present in the underlying neural network, $\f(\x_i; \w)$.
%For instance, if our augmentations are uniform over full 360 degree rotations, then $\finv(\x; \w)$ and $\pinv(\x; \w)$ will be invariant to rotations of the input [REF!!].
%The usual augmentations distributions used for image classification do take this form, and therefore do not induce full invariances. 
%Instead, they induced so-called ``soft'' invariance \citep{van2018learning}, meaning intuitively that $\finv(\x; \w)$ will vary less to perturbations of the input than $f(\x; \w)$, particularly when those perturbations are similar to the augmentation distribution.
%Finally, note that the averaging probabilities approach can also be understood as resulting from a ``noisy input'' model \citep{wenzel2020good} (Appendix~\ref{todo}).

The exact but intractable log-likelihoods, for averaging logits and averaging probabilities are
\begin{align}
  \Lprob^i\b{y_i; \w} &= \log \Pprob \bc{y_i}{\x_i, \w} \nonumber\\
  &= \log \E\sqb{\softmax_{y_i} \f(\x'_{i}; \w)},\label{eq:avgp_exact}\\
  \Llogits^i\b{y_i; \w} &= \log \Plogits\bc{y_i}{\x_i, \w} \nonumber \\
  &= \log \softmax_{y_i} \E\sqb{\f(\x'_{i}; \w)}.
\end{align}
We can form lower bounds on these quantities using $K$-sample estimators analogous to those in~\cite{burda2015importance}
\begin{align}
  \Lhprob^i\b{y_i; \w} &= 
       \log \sqb{\tfrac{1}{K} \tsum_{k=1}^K \softmax_{y_i} \f(\x'_{i; k}; \w)},\\
  \Lhlogits^i\b{y_i; \w} &= \log \softmax_{y_i} \sqb{\tfrac{1}{K} \tsum_{k=1}^K \f(\x'_{i; k}; \w)}.
\end{align}
To prove the lower bound for averaging probabilities, we first rewrite the expectation inside the logarithm of \myeqref{eq:avgp_exact} as the expectation of its average, over $K$ identically distributed random variables, $\x'_{i;k}$. We then take an approach familiar from variational inference \citep{jordan1999introduction} by applying Jensen's inequality to the (concave) logarithm function.   
\begin{align}
  \nonumber
  \Lprob^i\b{y_i; \w} &= \log \E\sqb{\tfrac{1}{K} \tsum_{k=1}^K \softmax_{y_i} \f(\x'_{i;k}; \w)}\\
   &\geq \E\sqb{\log \tfrac{1}{K} \tsum_{k=1}^K \softmax_{y_i} \f(\x'_{i;k}; \w)}\nonumber\\
   &= \E\sqb{\Lhprob^i\b{y_i; \w}}. \label{eq:bound_prob}
   \intertext{For averaging logits, we follow a similar method, noting that $\log \softmax_{y_i}$ is a concave function \citep{boyd2004convex} taking a vector of logits and returning a scalar log-probability for class $y_i$.  As such, we can again apply Jensen's inequality,}
  \nonumber
  \Llogits^i\b{y_i; \w} &= \log \softmax_{y_i} \E\sqb{\tfrac{1}{K} \tsum_{k=1}^K \f(\x'_{i;k}; \w)}\\
   &\geq \E\sqb{\log \softmax_{y_i} \tfrac{1}{K} \tsum_{k=1}^K \f(\x'_{i;k}; \w)} \nonumber\\
   &= \E\sqb{\Lhlogits^i\b{y_i; \w}}.\label{eq:bound_logits}
\end{align}
%These estimators are consistent in the sense that as $K\rightarrow \infty$ they become equal to the expectation.
%However, in practice we use finite and small $K$, raising the question of whether there are any guarantees (such as unbiasedness or lower-bounds) available for small $K$.
%Averaging logits was introduced in the Gaussian process setting, where such guarantees are available \citep{van2018learning}; however for neural networks we know of no method by which to obtain such guarantees and we suspect that this explains the community's reluctance to apply averaging logits to neural networks.
%In contrast, we can show that the finite sample estimator for averaging probabilities, $\Lhprob^i$, forms a lower bound by applying Jensen's inequality.
%First, note that for $\Lhprob^i$, the term inside the log is an unbiased estimator,
%\begin{align}
%  \nonumber
%  \E\sqb{\exp \Lhprob^i\b{y_i; \w}} &= \E\sqb{\tfrac{1}{K} \sum_{k=1}^K \softmax_{y_i} f(\x'_{i; k}; \w)} \\
%  &= \E\sqb{\softmax_{y_i} f(\x'_{i}; \w)} = \exp \Lprob^i\b{y_i; \w}
%\end{align}
%Taking the logarithm and applying Jensen's inequality we obtain,
%\begin{align}
%  \Lprob^i\b{y_i; \w} = \log \E\sqb{\exp \Lhprob^i\b{y_i; \w}} \geq \E\sqb{\Lhprob^i\b{y_i; \w}}.
%\end{align}
%so the expected value of $\Lhprob^i$ is a lower-bound on $\Lprob^i$, and by increasing $\Lhprob^i$, we can increase the true likelihood $\Lprob^i$.
%
%It is possible to eliminate the need for these approximations by considering the ``fixed-orbit'' setting...
Note that \citet{wenzel2020good} gave the single-sample averaging probabilities bound, but did not generalize it to tighter multi-sample bounds. We use ``multi-sample'' to describe bounds such as $\log \frac{1}{K}\tsum_k L(x_k)$ as in \citet{burda2015importance}. They are not to be confused with evaluating a single-sample bound with more Monte Carlo samples, $\frac{1}{K} \tsum_k \log L(x_k)$.

These bounds have a free parameter, $K$, raising the question of which values for $K$ are likely to be sensible.
We know that the bounds become tighter as $K$ increases \citep[e.g.][]{burda2015importance}, and eventually become exact as $K$ approaches infinity, suggesting that larger values of $K$ will better.
Remarkably, in variational inference (VI), practitioners frequently use a single-sample bound.
%Na\"ively, we would expect a single-sample estimate of an expectation to have high variance, and high-variance estimators give loose bounds and biased inferences when used with Jensen's inequality \citep{liao2018sharpening}.
%We do indeed find benefit in using $\Ktrain>1$ (Fig.~\ref{fig:class}), single-sample estimates are used frequently in variational inference \citep[VI;][]{jordan1999introduction}, which suggests they might be viable here too.
%That said, high variance and hence biased estimators are a problem in VI for the same reason, which is commonly mitigated by using $K>1$ \citep{burda2015importance,aitchison2018tensor}.
%That said, exactly the same issue (of high variance and bias from a single-sample bound) is encountered in VI, and a common strategy for reducing variance and tightening the bound is to use multiple samples  \citep[ $K>1$,][]{burda2015importance,aitchison2018tensor}.
However, VI incorporates a highly effective variance reduction strategy that is absent in our setting: an optimized variational approximate posterior (see Appendix~\ref{sec:app:vi_var}). 
%only reason they do something sensible is that in VI we optimizes a variational approximate posterior that is absent in our setting. 
%Optimizing the approximate posterior itself suppresses the variance and tightens the bound (see Appendix~\ref{sec:app:vi_var}).
In principle, similar variance reduction strategies exist in our setting, but would involve learning a separate variance-reducing augmentation distribution for each image, which is clearly impractical.
%In the absence of such a strategy, the only viable approach to reducing variance in the Jensen bound is to use multiple samples, though the exact number of samples required at test and train time is an empirical question.
%
%However, it is important to note that VI introduces a variational approximate posterior which is optimized, and one effect of this optimized approximate posterior is to minimize variance of the term inside the expectation.
%
%That said, it is important to note that VI is very different.
%To confirm that indeed, a single-sample estimator indeed represents a problematic approximation, note that 
%While one might think this choice would be immaterial or even beneficial, for our purposes it is very concerning as 
Indeed, in our setting, $K=1$ represents such a crude approximation that it collapses the differences between averaging probabilities, logits, and losses,
\begin{align}  
  \LhlogitsKone^i\b{y_i; \w} &=\LhprobKone^i\b{y_i; \w} = \Lhlogprob^i\b{y_i; \w}\label{eq:Ktrain=1}
\end{align}
which are all equal to $\log \softmax_{y_i} \f(\x'_{i; 1}; \w)$.
In contrast, we show empirically that $\Lhprob$, $\Lhlogits$ and $\Lhlogprob$ have significant performance differences when $K>1$ (Figs.~\ref{fig:class} and~\ref{fig:cold_posterior}). Note that we are free to use different numbers of samples at test and training time, $\Ktest$ and $\Ktrain$ respectively.

%This collapse is intuitive because no averaging actually occurs if we take $K=1$. 
% To get any difference between the methods, we need to actually take an average across two or more augmentations of the same underlying image.
%Thus, if we are interested in principled data augmentation priors, using one training sample is highly undesirable: it has such large approximation errors that it collapses the differences between averaging probabilities, averaging logits, and averaging losses.
%Indeed, the reason that \citet{wenzel2020good} considered only a single-sample estimator of the bound (i.e.\ $\Ktrain=1$) is precisely that it is equivalent to the standard data augmentation setup.% they used in the rest of the paper. %which is what they implemented in the rest of the paper).
%
%In general, we may use different values of $K$ at training and test time, $\Ktrain$ and $\Ktest$ respectively.

Finally, all of the above is for the usual ``full orbit'' setting, where there is a distribution over a very large, or even infinite number of possible augmentations.\footnote{We employ the term ``orbit'' from group theory and function invariance \citep{kondor2008group}, even though our augmentations do not necessarily form groups. In this work, it refers to the support of $\P(\x' | \x).$}
The full orbit setting necessitates the use of the bounds in \myeqref{eq:bound_prob} and~\myeqref{eq:bound_logits}.
Remarkably, if we consider an alternative ``finite orbit'' setting where only a small number of augmentations are available, we can \textit{exactly} evaluate the log-likelihood.
In the finite orbit setting, the distribution over augmented images, $\x_i'$, conditioned on the underlying unaugmented image, $\x_i$, can be written as,
\begin{align}
  \P\bc{\x_i'}{\x_i} &= \tfrac{1}{K} \tsum_{k=1}^K \delta\b{\x_i' - a_k(\x_i)},
\end{align}
where $\delta$ is the Dirac-delta, and $a_k$ is a function that applies the $k$th fixed augmentation.
In this setting, it is possible to exactly compute $\Llogits$ and $\Lprob$ by summing over the $K$ augmentations.
In practice, we choose the $K$ fixed augmentations by sampling them before training.
The finite orbit setting uses the same augmentations, and therefore the same number of augmentations, at test and train time: $\Ktrain=\Ktest=K$.

\section{RESULTS}
\label{sec:results}

\subsection{Principled DA in non-Bayesian networks}
\label{sec:sgd_results}
\newcommand{\augprob}{red}
\newcommand{\noaugprob}{green}
\newcommand{\auglogits}{orange}
\newcommand{\noauglogits}{blue}

\begin{figure*}
  \centering
  \includegraphics{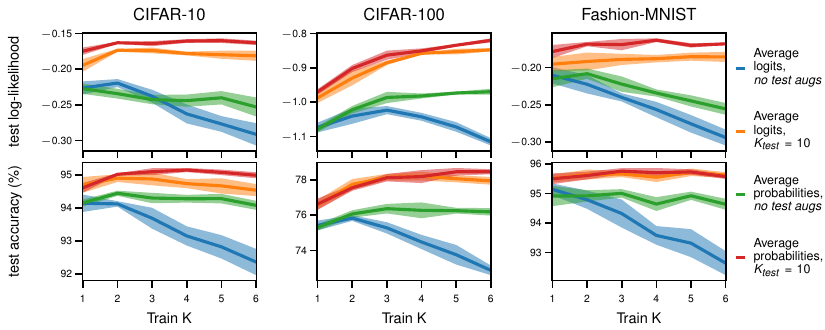}
  \caption{Comparison of averaging logits and probabilities for different values of $\Ktrain$, and using $\Ktest=10$ vs. using no test-time augmentations.
  Here, we use ResNet18 with SGD (i.e.\ no Bayesian inference).
  We use only full orbit to decouple $\Ktrain$ from $\Ktest$.
    \label{fig:class}
  }
  \vspace{-10pt}
\end{figure*}

We begin by comparing averaging logits and averaging probabilities in a non-Bayesian setting: SGD.
%While averaging logits (or its closely related cousin ``feature averaging'') is sometimes applied at test time \citep{he2015delving,benton2020learning,foster2020improving}, to our knowledge averaging logits is not used at train time in neural networks \citep[though see][which averages logits at train-time in GPs]{van2018learning}. Similarly, averaging probabilities has been used at test time \citep{krizhevsky2012imagenet}, but not at train time (to our knowledge).
Critically, higher values of $\Ktrain$ imply a larger computational cost per epoch, as each image is replicated and augmented $\Ktrain$ times before going through the network.
When assessing the benefit of averaging probabilities/logits for SGD training, we must therefore control for computational cost. We do this by training for $200/\Ktrain$ epochs.
Note that $\Ktrain=1$ with no test-time augmentation (i.e. \noaugprob{} and \noauglogits{} in Fig.~\ref{fig:class}) corresponds to the standard DA approach for both averaging logits and averaging probabilities \myeqref{eq:Ktrain=1}.
In this experiment, we consider only full orbit, which unlike finite orbit allows us to decouple $\Ktrain$ and $\Ktest$.

We trained ResNet18\footnote{\href{https://github.com/kuangliu/pytorch-cifar}{\nolinkurl{github.com/kuangliu/pytorch-cifar}}; MIT Licensed} on CIFAR-10, CIFAR-100 \citep{krizhevsky2009learning}\footnote{\href{https://cs.toronto.edu/~kriz/cifar.html}{\nolinkurl{cs.toronto.edu/~kriz/cifar.html}}} and FashionMNIST \citep{xiao2017fashion}\footnote{\href{https://github.com/zalandoresearch/fashion-mnist}{\nolinkurl{github.com/zalandoresearch/fashion-mnist}}; MIT Licensed} with a learning rate of 0.1, decayed to 0.01 three quarters of the way through training. We apply two augmentation transformations: 1. a random crop with padding of four pixels, on all borders and 2. a random horizontal flip with probability 0.5. The training runs took around 12 GPU-days on Nvidia 2080s.\footnote{Code available: \href{https://anonymous.4open.science/r/Augmentations-D513/}{\nolinkurl{anonymous.4open.science/r/Augmentations-D513/}}} %with averaging logits and averaging probabilities (Fig.~\ref{fig:class}).

In agreement with past work~\citep{lyle2020benefits}, we found that averaging over augmentations at test-time (\augprob{} and \auglogits{}) is better than using the test image without augmentation (\noaugprob{} and \noauglogits{}), with $\Ktrain=1$ corresponding to the standard DA procedure.
In addition, we show that improved performance with multiple test-time augmentations continues to hold for larger values of $\Ktrain$.
Thus, if sufficient compute is available at test-time, averaging across augmentations gives an easy method to improve the performance of a pre-trained network.

Importantly, we see some performance gains with higher values of $\Ktrain$ if we focus on the case with test augmentations, though they are somewhat inconsistent across datasets.
We see strong improvements for the hardest dataset (CIFAR-100), and smaller improvements that saturate at $\Ktrain=2$ for CIFAR-10.
For FashionMNIST, the picture is more mixed.
We suspect this is because we used a DA strategy tuned for CIFAR-10 and CIFAR-100, rather than FashionMNIST. %In particular, the augmentation includes a random horizontal flip which is unlikely to be a useful invariance for recognising digits in SVHN.
%These gains are smaller and saturate at $\Ktrain=2$ on easier datasets such as CIFAR-10, Fashion-MNIST and SVHN, but are very evident on the hardest dataset: CIFAR-100.

In addition, averaging probabilities seems to give somewhat better performance than averaging logits: compare averaging probabilities vs.\ logits both with test-time augmentation (\augprob{} vs.\ \auglogits{}) and without test-time augmentation (\noaugprob{} vs.\ \noauglogits{}).
%Further, note that the differences between averaging logits and averaging probabilities are smaller and somewhat consistent when using test-time augmentation (\augprob{} vs.\ \auglogits{}), but are much larger when using no test-time augmentation (\noaugprob{} vs.\ \noauglogits{}).
The performance differences are consistent in both comparisons, though smaller when test-time augmentation is applied.

\begin{figure}[t]
  \centering
  \includegraphics{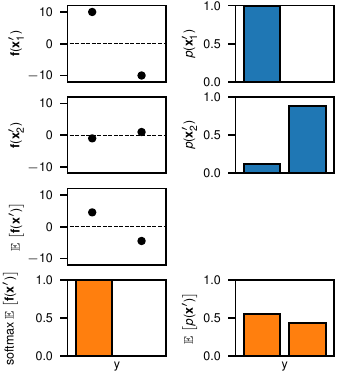}
  \caption{Example effect of averaging logits against averaging probabilities. $\x'_1$ and $\x'_2$ are two augmentations of the same image, $\f(\x'_1)$ and $\f(\x'_2)$ are logits outputted by a NN, and $p(\x'_1)$ and $p(\x'_2)$ are the probabilities corresponding  to these logits. The prediction derived from the averaged logits is much more certain than the average of the individual probabilities.
    \label{fig:avg_example}
  }
  \vspace{-10pt}
\end{figure}

Indeed, performance falls quite dramatically as $\Ktrain$ increases for averaging logits without test-time augmentation (\noauglogits{}).
This is an indication that averaging probabilities and logits might actually behave quite differently.
%perhaps the underlying neural network actually becomes less invariant as $K_\text{train}$ increases.
%This might occur because averaging over multiple augmentations introduces a degree of invariance in and of itself, so there is less need for the underlying neural network to be invariant.
%We would expect a less-invariant network to have reasonable performance with test-time augmentation (\auglogits{}) but to give worse performance without test-time augmentation (\noauglogits{}). From a statistical point of view, this is not problematic, as the same likelihood should be applied at test time as at training time, which prescribes augmentation at test time. From a practical point of view however, this behaviour is problematic, as we may want to use a small $\Ktest$ for computational efficiency at test time.
%% While this can perhaps be resolved by just using $\Ktrain=1$, it is extremely problematic in our statistical framework because as $\Ktrain$ increases, we get closer to the correct log-likelihood.
%In contrast, this does not occur with averaging probabilities: performance is much more constant as $\Ktrain$ increases, even when evaluating without test-time augmentation (\noaugprob{}).
%
%
To understand how these differences might arise, consider the effect of averaging on the NN function itself. Both schemes can be justified by using averaging to increase invariance to the augmentation transformations (Sec.~\ref{sec:tighter}). Averaging probabilities, however, also forces the NN function itself to become invariant. If different augmentations produce different predictions, then the resulting averaged prediction will be more uncertain, which is penalized by the likelihood on the training points. This effect is much weaker when averaging logits.
Consider an extreme example, as illustrated in Fig.~\ref{fig:avg_example}. It is a two-class classification problem with two augmentations, $\x'_1$ and $\x'_2$, of the same image with logits, $\f(\x'_1) = (10, -10)$ and $\f(\x'_2) = (-1, 1)$.
Averaging logits gives us $\E\sqb{\f(\x')} = (4.5, -4.5)$, and applying the softmax, we very confidently predict the first class.
In contrast, if we use averaging probabilities, then the first augmentation almost certainly predicts the first class $p(\x'_1) \approx (1, 0)$ and the second augmentation almost certainly predicts the second class, $p(\x'_2) \approx (0, 1)$, so when we average them we obtain $\E\sqb{p(\x')} \approx (0.5, 0.5)$, which indicates a high degree of uncertainty.

\subsection{Bayesian neural networks and the cold posterior effect}
\label{sec:res:cold_post}

\begin{figure*}
  \centering
  \includegraphics[width=0.9\textwidth]{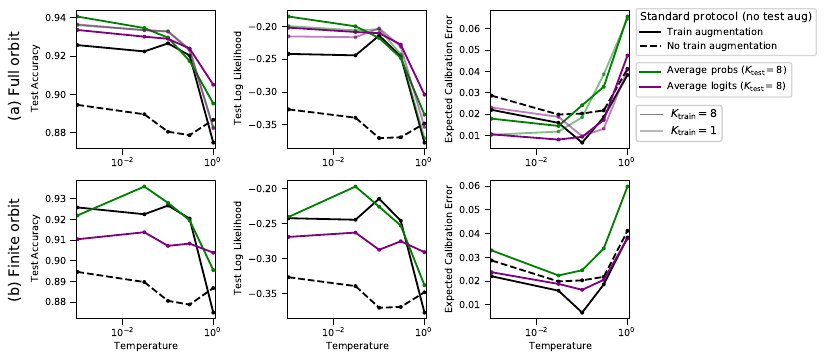}
  \caption{The cold posterior effect for different DA setups when running GGMC with ResNet20 on CIFAR-10. Without DA, there is a minimal CPE. Most other configurations show significant improvement for $T<1$, with the exception of averaging the logits over a finite orbit. Averages computed with $\Ktrain=8$ and $\Ktest=8$.
  }
  \label{fig:cold_posterior}
  \vspace{-10pt}
\end{figure*}

Next, we ask a very different question: how is the CPE influenced when DA is incorporated into the model in a principled way?
To this end, we use a different experimental setup.
In particular, we take the code\footnote{\href{https://github.com/ratschlab/bnn_priors}{\nolinkurl{github.com/ratschlab/bnn_priors}}; MIT Licensed} and networks from \citet{fortuin2021bayesian,fortuin2021bnnpriors} and mirror their experimental setup as closely as possible.
This code combines a cyclical learning rate schedule \citep{zhang2019cyclical}, a gradient-guided Monte Carlo (GGMC) scheme \citep{guided-mcmc}, and the preconditioning and convergence diagnostics from \citet{wenzel2020good}. The DA transformations are the same as those described in Sec.~\ref{sec:sgd_results}. 
Following \citet{fortuin2021bayesian}, we ran 60 cycles with 50 epochs in each cycle. 
We recorded one sample at the end of each of the last five epochs of a cycle, giving 300 samples total. 
Importantly, to allow for running many sampling epochs in these experiments, we follow \citet{fortuin2021bayesian} in using the ResNet20 architecture from \citet{wenzel2020good}, which has far fewer channels than the ResNet18 used in Sec.~\ref{sec:sgd_results} (i.e.\ 32 channels for the first block up to 128 in the last block compared to 64 channels up to 512 \citep{he2016deep}).
As such, SGD in this network performs poorly compared with that in Sec.~\ref{sec:sgd_results} ($\sim 92\%$ \citep{wenzel2020good} vs. $\sim 95\%$ \citep{he2016identity}). The experiments took around 60 GPU-days on Nvidia RTX6000s\footnote{Code available: \href{https://anonymous.4open.science/r/bayesian-data-aug/experiments/bayes_data_aug/README.md}{\nolinkurl{anonymous.4open.science/r/bayesian-data-aug/experiments/bayes_data_aug/README.md}}}.% For all augmentation and averaging configurations, we found the posterior predictive at some non-zero temperature to be more accurate than its SGD equivalent (see Appendix~\ref{app:sgd_cold_posterior}).

The results are presented in Fig.~\ref{fig:cold_posterior}. We replicate the finding that the CPE is largely absent without DA (dashed black line), and is present in the standard setup with DA at training time ($\Ktrain=1$) but without augmentation at test time (solid black).
Further, we show that the CPE persists with principled DA likelihoods: averaging logits with full orbit (purple, top row), and averaging probabilities with finite and full orbits (green).
The best method overall appears to be averaging probabilities with a full orbit (dark green line, top row) at $T=0.001$, though averaging logits (dark purple lines) is better at $T=1$.

Surprisingly, the CPE is absent in one particular setting: averaging logits with a finite orbit (dashed purple line).
However, the relevance of this is unclear, as it is clearly the worst performing of all our approaches by quite some margin.
Indeed, remember that the arguments for the optimality of Bayesian inference apply only in the case that the model is well-specified \citep{kolmogorov1950foundations,savage1954foundations,jaynes2003probability}.
However, the poor performance of averaging logits with a finite orbit indicates that it is likely to be the wrong model, while other settings are likely to be closer to the true model.
In that case, the presence or absence of the CPE in the wrong model (averaging logits with a finite orbit) is immaterial to our understanding of the CPE in the right model.
See also Sec.~\ref{sec:conclusions}; note that this argument could not be made if there was a model without the CPE with performance equal to or better than the other models.

The CPE was originally discovered in \citet{wenzel2020good} when assessing test accuracy and log-likelihood --- they did not consider other measures of distribution calibration like ECE.
Indeed, later work on the cold posterior effect found that measures such as ECE are far more complex and usually do not agree with test accuracy and log-likelihood \citep{fortuin2021bnnpriors}.
It is therefore difficult to interpret the differences between test log-likelihod and ECE, especially if we remember that test log-likelihood is itself a proper scoring rule \citep{gneiting2007strictly}, and therefore it does capture one possible notion of calibration.
In particular, test log-likelihood heavily penalizes an event assessed as low probability actually happening, e.g.\ if our classifier predicts a probability of $0.001\%$, while the actually happens even $0.1\%$ of the time.
In contrast, ECE considers the absolute difference in probability, so it far more heavily penalises e.g.\ a predicted probability of $40\%$ while the event actually happens $60\%$ of the time.
Needless to say, the most appropriate measure of calibration will depend heavily on the domain, with log-likelihood being appropriate for low-probability but high risk events: in the example considered here of $0.001\%$ vs $0.1\%$, the factor of 100 error in the estimated probability could be catastrophic, despite representing a tiny absolute error of only $0.1\% - 0.001\% = 0.099\%$.

%In the finite orbit setting, it is evident that performance is much better with averaging probabilities.
%Surprisingly, the cold posterior effect is much smaller for averaging logits with finite orbit (dashed purple line), which demonstrated best performance at $T=1$ (close to  averaged logits over full orbit; solid purple line).
%However, the relevance of this is unclear, as the reduction in the cold posterior effect arises primarily because of worse low-temperature performance.
%However, the performance at lower temperatures is far worse than other approaches such as those based on averaging probabilities.

%This suggests that the 
%However, the relevance of this finding is debatable: if we believe that $T=1$ is the Bayesian solution then this result is highly relevant.
%However, Bayesian inference in the correct model should would have $T=1$ as the optimal temperature, and give optimal performance, which it clearly does not.

The usefulness of these results is contingent on understanding whether we are indeed accurately approximating the posterior.
To check this, we computed the kinetic temperature~\citep{leimkuhler2015canonical}, which estimates the temperature of a given parameter in the Langevin dynamics simulation from the norm of its momentum. 
In expectation, the kinetic temperature estimator should be equal to the desired temperature, $T$.
The results (Appendix~\ref{sec:app:kinetic_temp}) indicate that all the samplers run at their desired temperature, a result that is consistent with accurate posterior sampling.

\begin{figure}[t]
  \centering
  \includegraphics[width=0.43\textwidth]{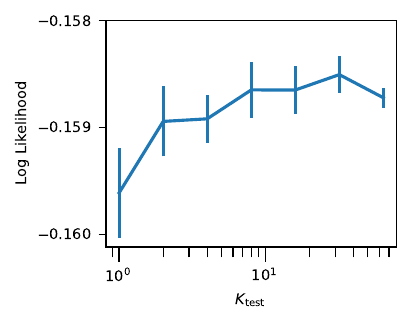}
  \caption{The effect of $\Ktest$ on the log-likelihood bound for a test batch (size 512) of CIFAR-10. Values shown for ResNet20 BNN trained and tested with $\Lhprob$ ($\Ktrain=8$ and $T=0.001$). Error bars cover two standard errors above/below mean for DA sampling with different seeds. Sixty seeds used for $\Ktest=\{1,2\}$, thirty for $\Ktest=4$ and five for all other $\Ktest$.}
  \label{fig:loglik_vs_Ktest}
  \vspace*{-10pt}
\end{figure}

As discussed in Sec.~\ref{sec:meth}, increasing $K$ tightens our log-likelihood bounds. However, increasing $K$ also incurs greater computational cost. It is natural to question which value of $K$ is a good trade-off. To answer this, we explore how the log-likelihood of test data under a trained model varies with $\Ktest$. As expected, the results (Fig.~\ref{fig:loglik_vs_Ktest}) show the log-likelihood increases with $\Ktest$, with even $K=2$ being a significant improvement over $K=1$ (standard DA). Further, the curve plateaus, suggesting that for CIFAR-10, there is little benefit of using $K>8$.

\section{RELATED WORK\label{sec:related_work}}

Past work introduced generative models which average probabilities \citep{wenzel2020good}.
However, this work did not consider the tight multi-sample bounds developed here, or the finite orbit setting which allows us to evaluate the exact likelihood.
This left open the possibility raised by \citet{izmailov2021bayesian} that the CPE was an artifact of standard DA resulting in an invalid likelihood.
In contrast, we considered exact likelihoods in the finite orbit setting, and tight multi-sample lower bounds in the full orbit setting.
As the CPE persists when using our exact likelihoods or tight lower bounds, we can exclude the possibility that the CPE is an artifact of DA giving a ``randomly perturbed log-likelihood''.
Other work has introduced a log-likelihood estimator for averaging GP logits \citep{van2018learning}. However, the method only works for a quadratic log-likelihood and thus necessitates P\'{o}lya-Gamma approximations for classification. Further, the work did not consider BNNs or the connection to the CPE.

There is a small but growing body of work that considers averaging over multiple augmentations at training time \citep{hoffer2019augment,berman2019multigrain,choi2019faster,benton2020learning,lyle2020benefits,touvron2021going}. 
The issue is still highly topical, with important contemporaneous work \citep{fort2021drawing}.
However, this work was not done within a Bayesian framework (e.g.\ by using stochastic gradient Langevin dynamics (SGLD) or a similar inference algorithm), did not show that averaging across multiple training augmentations gives a tight multi-sample bound on the log-likelihood of a principled model, did not consider the finite-orbit setting where the log-likelihood can be computed exactly, and did not consider the interaction with the CPE.
In addition, much of this work uses averaging losses \citep{benton2020learning,touvron2021going,fort2021drawing} which is equivalent to using a loose single-sample bound on the log-likelihoods.
%Past \citep{lyle2020benefits} and contemporaneous work \citep{fort2021drawing}, suggests that in agreement with our results in Fig.~\ref{fig:class}, using multiple training augmentations improves generalization in large neural networks.
%However, \citep{lyle2020benefits}   In addition, \citep{fort2021drawing} considered only averaging losses. % within a principled Bayesian framework as they used averaging losses, and they did not assess any connection to the cold posterior effect. 
%Similarly, \citet{hoffer2019augment} and~\citet{berman2019multigrain} averaged losses for improving the generalization of models trained with large batches, and \cite{choi2019faster} did so for better hardware utilization. \cite{benton2020learning} averaged losses in order to learn invariances in neural networks.
%In this case, there is no difference in the expected training objective for averaging one sample vs.\ multiple samples, so \citet{benton2020learning} used a single sample.
%Last, the averaging of losses is applied in the context of vision transformers by \cite{touvron2021going}. 
Finally, the idea of averaging at test-time is more common and has been practiced for longer \citep[e.g.][]{krizhevsky2012imagenet,simonyan2014very,he2015delving,szegedy2015going,foster2020improving}.
A considerable body of past work on BNNs uses DA, both with variational inference \citep{blundell2015weight} and SGLD \citep[e.g.][]{zhang2018noisy,zhang2019cyclical,osawa2019practical,fortuin2021bayesian,immer2021scalable}.
However, as discussed in Sec.~\ref{sec:background} (Background), these methods simply substitute non-augmented for augmented data and thus do not use a valid log-likelihood.
In contrast, we incorporated DA into the probabilistic generative model, and thus are able to give valid log-likelihoods based on averaging logits or averaging probabilities.

%\section{Conclusions}
%We considered two principled probabilistic generative models incorporating DA. We found that they give improved performance in standard SGD settings with a fixed compute budget.
%In addition, we looked at their interaction with the cold posterior effect.
%The cold posterior effect did not go away when using principled DA, suggesting that the question of dataset size in DA may not provide an explanation of the cold posterior effect.
%Instead, these results are consistent with an alternative theory of the cold posterior effect, which introduces a probabilistic generative model describing data curation \citep{aitchison2020cold}.
%No particular negative social impacts are anticipated as this is largely theoretical work.

\section{IMPLICATIONS FOR THE CPE}
\label{sec:conclusions}
We have shown that the CPE persists even when using models incorporating principled DA, and in agreement with past work \citep{wenzel2020good,fortuin2021bayesian,izmailov2021bayesian}, we show that the CPE disappears without DA.

What do these results imply for the origin of the CPE?
First, our models in principle have a clean log-likelihood which can be evaluated exactly in the finite orbit setting, or which we estimate using tight multi-sample bounds in the full orbit setting.
Thus, we can no longer dismiss the CPE as an artifact arising from DA giving a ``randomly perturbed log-likelihood [which] does not have a clean interpretation as a valid likelihood function''.

Indeed, it is worth stepping back and considering the original motivation for studying the CPE, namely that if we have the correct model, then Bayesian inference with T=1 should give optimal performance \citep{kolmogorov1950foundations,savage1954foundations,jaynes2003probability,wenzel2020good}.
Critically, we need the right model for us to expect optimal performance at T=1.
We now have two classes of model, with DA and without DA, so which is right(er)?
Given the significant and widely recognised performance benefits of DA, it seems very likely that the ``right'' model would include some form of DA. %, and that the model without DA is ``wrong(er)''.
If the model with DA is right(er), and that model displays the CPE, then the CPE still demands an explanation, and the presence or absence of the CPE in the wrong model without DA is immaterial.
As such, the presence of the CPE in models with DA remains an important problem, and is likely to be caused by one of the two other explanations discussed in Sec.~\ref{sec:intro} (Introduction): either data curation \citep{aitchison2020cold} or prior misspecification \citep{wenzel2020good,fortuin2021bayesian}.
Indeed, we would tentatively suggest the opposite of \citet{izmailov2021bayesian}: that it is in reality the \textit{lack} of a CPE without DA that is an artifact of using the wrong model (i.e.\ without DA).

Finally, note that the CPE is not always observed, e.g.\ in language modelling \citep{izmailov2021bayesian}.
This is absolutely expected as the data-curation explanation of \citet{aitchison2020cold} only implies CPE in fairly restricted settings; i.e.\ \textit{only} in the case of reasonably accurate approximate posterior inference, such as SGLD, in a BNN where the data has been curated by excluding datapoints with an ambiguous class-label.
Thus, \citet{aitchison2020cold} does not lead us to expect the CPE e.g. in latent variable models, in regression settings (where you typically do not curate data), or in hybrid models where we perform Bayesian inference over only a small subset of parameters.

%\pagebreak

%\section{Discussion}
%
%
%%Intuitively, we can think of averaging logits as willing to give a confident answer if it finds one augmentation under which the image-class is obvious.
%%Thus,  underlying network can be non-invariant
%%In contrast averaging probabilities is only willing to give a confident answer if the classification is reasonably
%
%Our work has important implications for a potential explanation of the cold posterior effect.
%There are ... potential explanations.
%
%
%Second, ...

% \bibliographystyle{icml2021}
\bibliography{refs}

\newpage
\onecolumn
\thispagestyle{empty}
\aistatstitlenovskip{Supplementary material for data augmentation in Bayesian neural networks and the cold posterior effect}
% \vspace*{-120pt}

\appendix

\section{AVERAGING LOSSES EMERGES WHEN USING DA IN VI AND SGLD\label{sec:app:vi_sgld}}

There are two particularly important algorithms for doing Bayesian inference in neural networks: stochastic gradient Langevin dynamics \citep[SGLD;][]{welling2011bayesian} and variational inference \citep[VI;][]{blundell2015weight}.
In SGLD without DA, we draw samples from the posterior over weights by following gradient of the log-probability with added noise,
\begin{align}
% \begin{split}
  \b{\Delta \w}_\text{noaug} =\frac{\epsilon}{2} \nabla_{\w} \Bigg[\log \P\b{\w} + \sum_{i=1}^N \log \P_\text{noaug}\bc{y_i}{\x_i, \w} \Bigg]  + \sqrt{\epsilon} \; \boldsymbol{\eta}
  \label{eq:sgld}
  %&= \frac{\epsilon}{2} \nabla_{\w} \sqb{\P\b{\w} + \sum_{i=1}^N \log \softmax_{y_i}\b{f(\x_i; \w)}}  + \sqrt{\epsilon} \; \boldsymbol{\eta}
%  \end{split}
\end{align}
where $\boldsymbol{\eta}$ is standard Gaussian IID noise, and for simplicity we give the expression for full-batch Langevin dynamics rather than minibatched SGLD (they do not differ for the purposes of reasoning about DA). 
Likewise the variational inference objective is,
\begin{align}
  \label{eq:elbo}
  \text{ELBO}_\text{noaug} = \E_{\Q\b{\w}}\Bigg[\log\P\b{\w} +\sum_{i=1}^N \log \P_\text{noaug}\bc{y_i}{\x_i, \w} 
   -\log \Q\b{\w}\Bigg]
\end{align}
where $\Q\b{\w}$ is the variational approximate posterior learned by optimizing this objective.
%The typical approach to incorporating DA in these settings is to replace the unaugmented image, $\x_i$, with an augmented image, $\x'_i$, in Eq.~\eqref{eq:sgld} and \eqref{eq:elbo}.
%\begin{align}
%  \E\sqb{\b{\Delta \w}_\text{aug}} %&= \frac{\epsilon}{2} \nabla_{\w} \sqb{\P\b{\w} + \sum_{i=1}^N \E\sqb{\log \softmax_{y_i}\b{f(\x_i'; \w)}}}  + \sqrt{\epsilon} \; \boldsymbol{\eta}\\
%  &= \frac{\epsilon}{2} \nabla_{\w} \sqb{\P\b{\w} + \sum_{i=1}^N \Llogprob^i\b{y_i; \w} }  + \sqrt{\epsilon} \; \boldsymbol{\eta},\\
%  \text{ELBO}_\text{aug} &= \E_{\Q\b{\w}}\sqb{\P\b{\w} + \sum_{i=1}^N \Llogprob^i\b{y_i; \w} - \log \Q\b{\w}}.
%\end{align}
To understand the overall effect of this approach to DA, we replace $\log \P_\text{noaug}\bc{y_i}{\x_i, \w}$ with the log-softmax using \myeqref{eq:def:Lnoaug}.
Then, we consider the expected update to the weights, averaging over the augmented images, $\x_i'$ conditioned on the underlying unaugmented images, $\x_i$,
\begin{align}
  \E\sqb{\b{\Delta \w}_\text{aug}} %&= \frac{\epsilon}{2} \nabla_{\w} \sqb{\P\b{\w} + \sum_{i=1}^N \E\sqb{\log \softmax_{y_i}\b{f(\x_i'; \w)}}}  + \sqrt{\epsilon} \; \boldsymbol{\eta}\\
  &= \frac{\epsilon}{2} \nabla_{\w} \Bigg[\log\P\b{\w} +\sum_{i=1}^N \Llogprob^i\b{y_i; \w} \Bigg]  + \sqrt{\epsilon} \; \boldsymbol{\eta},\\
  \text{ELBO}_\text{aug} &= \E_{\Q\b{\w}}\Bigg[\log\P\b{\w}+ \sum_{i=1}^N \Llogprob^i\b{y_i; \w} - \log \Q\b{\w}\Bigg].
\end{align}
In both cases, this ultimately replaces $\log \P_\text{noaug}\bc{y_i}{\x_i, \w}$ with $\Llogprob^i\b{y_i; \w}$, which as discussed in Sec.~\ref{sec:meth} is not a valid log-likelihood.

%\section{Principles for Incorporating DA in Probabilistic Models}

\section{THE APPROXIMATE POSTERIOR IN VI REDUCES VARIANCE}
\label{sec:app:vi_var}

Here, we derive the ELBO using Jensen's inequality; we take $x$ to be the data and $z$ to be a latent variable.
Our goal is to compute the model evidence, $\P\b{x}$, by integrating out $z$,
\begin{align}
  \P\b{x} &= \int dz \P\bc{x}{z} \P\b{z} =\int dz \P\b{x,z}
\end{align}
where $\P\b{z}$ is the prior, $\P\bc{x}{z}$ is the likelihood and $\P\b{x, z}$ is the joint.
We introduce an approximate posterior, $\Q\b{z}$, and rewrite the integral as an expectation over that approximate posterior and apply Jensen's inequality,
\begin{align}
  \log \P\b{x} &= \log \int dz \Q\b{z} \frac{\P\b{x,z}}{\Q\b{z}} \\
  &= \log \E_{\Q\b{z}}\sqb{\frac{\P\b{x,z}}{\Q\b{z}}}\geq\E_{\Q\b{z}}\sqb{\log \frac{\P\b{x,z}}{\Q\b{z}}}.
\end{align}
Now it is evident that the tightness of the bound is controlled by the variance of $\P\b{x, z} / \Q\b{z}$.
Critically, if $\Q\b{z}$ matches the true posterior,
\begin{align}
  \Q\b{z} &= \P\bc{z}{x} \propto \P\b{x, z}
\end{align}
then $\P\b{x, z} / \Q\b{z}$ is constant (zero variance) and the bound is tight.

% \pagebreak
\section{KINETIC TEMPERATURE DIAGNOSTIC RESULTS}
\label{sec:app:kinetic_temp}
\begin{figure*}[h!]
  \centering
  \includegraphics{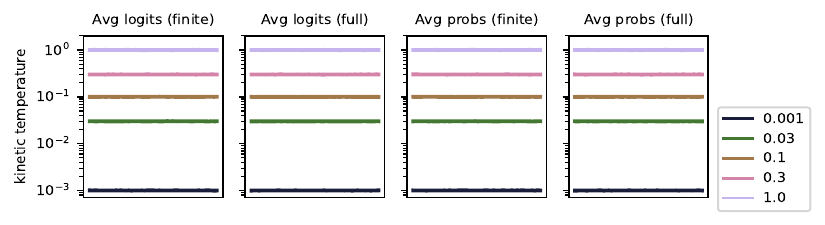}
  \caption{The evolution of the kinetic temperature diagnostic~\citep{leimkuhler2015canonical} throughout inference. Good agreement between the diagnostic temperature and intended temperature (in legend) suggests accurate inference.}
  \label{fig:kinetic_temp}
\end{figure*}
\FloatBarrier

\section{GENERALIZATION OUTSIDE OF CLASSIFICATION}
\label{sec:app:generalization}
We may be interested in generalizing the averaging logits and averaging probabilities ideas outside  classification.
For averaging probabilities, we use,
\begin{align}
  \P\b{y_i| \x_i} &= \int d \x'_i \P\b{y_i| \x_i'} \P\b{\x_i'| \x_i} = \E\sqb{\P\b{y_i| \x_i'}}
  \label{eq:avg_p}
\end{align}
Intuitively, each augmentation forms one component of a (potentially infinite) mixture model over the outputs, $y_i$.
Importantly, this expression makes no assumption about the support of distributions over $y_i$, so $y_i$ could be a finite set (classification), real-valued (regression), or anything else (a string, a graph, etc.) 
Note that directly applying a multi-sample estimator to (the logarithm of)~\myeqref{eq:avg_p} gives us a log-likelihood lower bound as in~\myeqref{eq:bound_prob}.

To generalize averaging logits, consider a situation where a distribution over an arbitrary $y_i$ is parameterized by a vector, $\f_i$ output by a neural network,
\begin{align}
  \P\b{y_i| \x} &= \pi\b{y_i; \f_i}.
\end{align}
In the standard case with no augmentation, we would take $\f_i=\f(\x_i; \w)$, (where we take $\f_i$ as the specific vector for input $i$, and $\f(\cdot; \cdot)$ as a function represented by a neural network, that takes an image and weights and returns a vector).
In the case with augmentation, we can average neural network outputs across different augmentations,
\begin{align}
  \f_{i} &= \int d \x'_i \P\b{\x_i'| \x_i} \f(\x_i'; \w) = \E\sqb{\f(\x_i'; \w)}.
  \label{eq:avg_f}
\end{align}
Note that in this case we need additional conditions for the multi-sample estimator to form a lower bound.
In particular, we need $\log \pi\b{y_i; \f_i}$ to be concave when treated as a function of $\f_i$ for a fixed $y_i$.% We explore the properties and justifications of both~\myeqref{eq:avg_p} and~\myeqref{eq:avg_f} in Appendix~\ref{sec:app:perspectives}.

\section{PERSPECTIVES OF PROBABILISTIC DATA AUGMENTATION}
\label{sec:app:perspectives}
Here we explore in depth the two general approaches to probabilistic data augmentation~\myeqref{eq:avg_p} and~\myeqref{eq:avg_f}. We discuss their justifications in Sec.~\ref{sec:app:invariance} and~\ref{sec:app:noisy_input}, and compare their properties in Sec.~\ref{sec:app:behaviour}.

\subsection{Invariance construction}
\label{sec:app:invariance}
In the main text, we suggest two ways of incorporating data augmentation: 1) by averaging logits output by the neural network, and 2) by averaging the predicted probabilities. In classification, both of these methods can be justified by attempting to create a prediction that is more invariant to the transformations in the data augmentation.

When averaging logits, we aim to make the neural network mapping $\f: \mathbb{R}^D \to \mathbb{R}^C$ more invariant by averaging the outputs as in~\myeqref{eq:avg_f}.
% \begin{equation}
%     \label{eq:invavgfunc}
%     \finv(\x; \w) = \E\sqb{\f(\x'; \w)} = \int \f(\x'; \w) \P(\x'|\x) \mathrm{d}\x' \,.
% \end{equation}
This construction influences only the regression function, and so has a similar effect to changing the neural network architecture or changing the prior on the functions $\f(\cdot)$ in the Bayesian case \citep{van2018learning}. Since only the outputs are affected, this can be directly applied to any likelihood that depends only on an evaluation of the function, i.e.~any likelihood which can be written as $\P(y_i | \f_i)$.

In the case of averaging the probabilities, we can consider the model to be learning a mapping from image inputs to probability vectors $\p: \mathbb{R}^D \to \mathbb{P^C}$. We can make this mapping more invariant in the same way:
\begin{align}
    \label{eq:invavgprob}
    \pinv(\x_i;\w)&= \E\sqb{\softmax \left(\f(\x'_i; \w)\right)} \nonumber\\
    &= \int \softmax \left(\f(\x'_i; \w)\right) \P(\x'_i|\x_i) \mathrm d\x'_i\,.
\end{align}
The straightforward generalization of this construction would be to replace the softmax with the appropriate likelihood (see general case in~\myeqref{eq:avg_p}). When considering likelihoods other than softmax classification (e.g.~Gaussian likelihoods for regression), stronger differences between these constructions emerge in both behaviour and justification. We investigate further in  Appendix~\ref{sec:app:behaviour}.

\subsection{Noisy-input model}
\label{sec:app:noisy_input}
As stated above, we can generalize averaging the classification probabilities by replacing the softmax with the appropriate likelihood as in~\myeqref{eq:avg_p}.
% \begin{align}
%     p_\text{inv}(y_n|\x_n, \w) = \int p(y_n|\w, \x_n') p_\text{aug}(\x_n'|\x_n) \mathrm{d}\x_n' \,.
%     \label{eq:noisy_input}
% \end{align}
This modified likelihood, which incorporates data augmentation, was also discussed in \citet[Appendix~K]{wenzel2020good} and is a (potentially continuous) mixture model on the observation $y_n$, where each augmentation introduces a mixture component.  This is as a \emph{noisy-input} model \citep{girard2003learning,mchutchon2011gaussian,damianou2016variational} where the input $\x_i$ is corrupted via the augmentation distribution.

\subsection{Model comparison}
\label{sec:app:behaviour}
The forms of the invariance construction~\myeqref{eq:avg_f} and the noisy-input model~\myeqref{eq:avg_p} imply a difference of purpose. In using the invariance construction, we seek a regression function with the specified symmetry, which is consistent with the data according to the likelihood function $\P\lrbracket{(}{)}{y_i | \f_i}$. Conversely, with the noisy-input model~\myeqref{eq:avg_p} we aim to find a function which gives rise to an invariant likelihood, consistent with the observed outputs for inputs randomly perturbed by $\P(\x'|\x)$. The role of $\x'$ is different in each case. In the noisy-input model, $\x'$ is a latent variable on which we could, in principle, do inference (with e.g. an amortized variational approach). While in the invariance construction, we integrate over $\x'$ to parameterize $\f(\x;\w)$.

We now compare the behaviours of the invariance and noisy-input constructions. We will see that they result in quite different posteriors.

In the main text, we compared the empirical performance of averaging probabilities and averaging logits for BNN classification (see Figs.~\ref{fig:class} and~\ref{fig:cold_posterior}). However, as the invariance perspective justifies both averaging logits and probabilities, this comparison does not clearly distinguish between the noisy-input and invariance viewpoints. Further, we are interested not only in predictive performance but also in understanding how each construction behaves. With this in mind, we investigate the models with an illustrative example, where we can both integrate over the orbit and do inference in closed form.

\begin{figure*}[t]
    \begin{subfigure}{0.38\textwidth}
        \adjincludegraphics[trim={0 0 {.6\width} 0},clip,width=\textwidth]{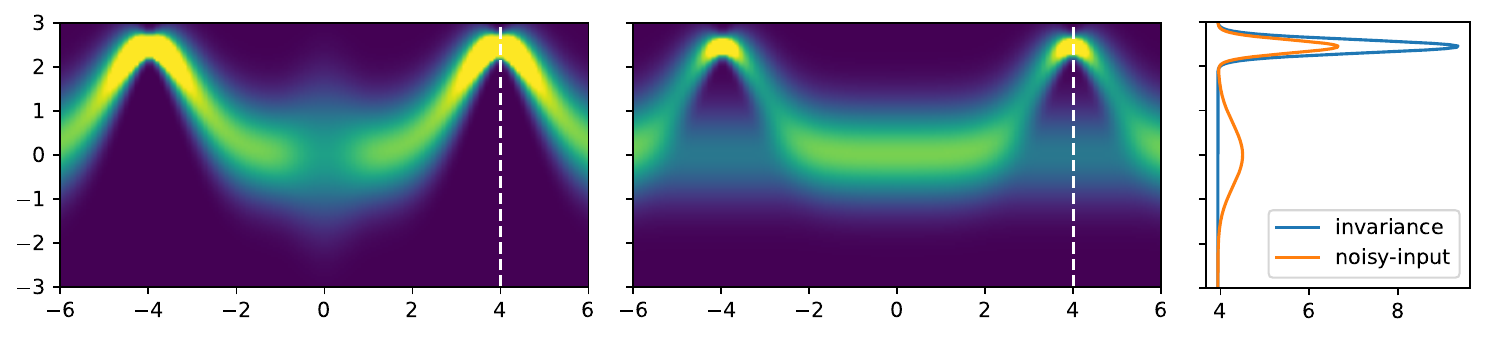}
        \caption{Invariant GP}
    \end{subfigure}%
    \begin{subfigure}{0.38\textwidth}
        \adjincludegraphics[trim={{.4\width} 0 {.2\width} 0},clip,width=\textwidth]{figs/app_general_lhoods/posterior_densities.pdf}
      \caption{Noisy-input GP}
        \label{subfig:noisy_input_posterior}
    \end{subfigure}%
    \begin{subfigure}{0.19\textwidth}
      \adjincludegraphics[trim={{.8\width} 0 0 0},clip,width=\textwidth]{figs/app_general_lhoods/posterior_densities.pdf}
      \caption{$p(y|x=4,\mathcal{D})$}
      \label{subfig:noisy_input_vs_invariance_marginals}
    \end{subfigure}
    \caption{Posterior densities for the model constructions for a single observation at $x_1=-4,y_1=2.5$.}
    \label{fig:noisy_input_vs_invariance_densities}
\end{figure*}

We consider Gaussian process (GP) regression with a one-dimensional input and data augmentation which enforces symmetry about $x=0$, i.e. $\P(x'|x)=\frac{1}{2}\left(\delta(x'-x)+\delta(x'+x)\right)$. From \Citet{van2018learning}, the invariance view can be expressed in the kernel of the GP:
\begin{align}
    g&\sim\mathcal{GP}(\mathbf{0},k_{\text{base}})\\
   f(x)&=g(x)+g(-x)\\
    \implies f&\sim\mathcal{GP}(\mathbf{0},k_\text{inv}),\\
       \text{where} \hspace{15pt} k_\text{inv}(x_i,x_j) &= \sum_{c_i\in\{-1,1\}} \sum_{c_j\in\{-1,1\}}k_{\text{base}}(c_i x_i,c_j x_j).
\intertext{
We then follow standard GP inference to find the posterior over invariant functions. Note that unlike \Citet{van2018learning}, we are not concerned with learning invariances here.\endgraf
The noisy-input model for this case is}
    \P(\x,\y,\f)&= \P(\f)\prod_{i=1}^N \int \P(y_i|f(x_i'))\P(x_i'|x_i)dx_i'\label{eq:noisy_input_joint}\\
    \P(y_i|f(x_i'))&= \mathcal{N}\left(y_i;f(x_i'),\sigma^2\right)\\
    f&\sim \mathcal{GP}(0, k).
\intertext{
Given a single observation $(x_1,y_1)$, the noisy-input posterior is} 
    \P(f|x_1,y_1)&= \frac{1}{Z}\P(f(x_1),x_1,y_1)\\
    &=\frac{1}{2Z}\P\lrbracket{(}{)}{f(x_1)}\sqb{\P(y_1|f(x_1))+\P(y_1|f(-x_1))}\\
    &= \frac{1}{2}\sqb{\P(f|x_1,y_1)+\P(f|-x_1,y_1)},
\end{align}
a mixture of GP posteriors, with two components (one for each point in the orbit).

How do these posteriors compare? For an observation at $(x_1=-4,y_1=2.5)$ we plot the posterior predictive densities in Fig.~\ref{fig:noisy_input_vs_invariance_densities}. Both posteriors are symmetric around $x=0$ as we expect, however the noisy-input model is bimodal in the regions surrounding $x=4$ and $x=-4$, where the invariance posterior has unimodal density concentrated around the observed $y$ value of 2.5. The difference is clear in Fig.~\ref{subfig:noisy_input_vs_invariance_marginals}, which shows the marginal predictive densities at $x=4$.

\begin{figure*}[t]
    \centering
    \includegraphics[width=\textwidth]{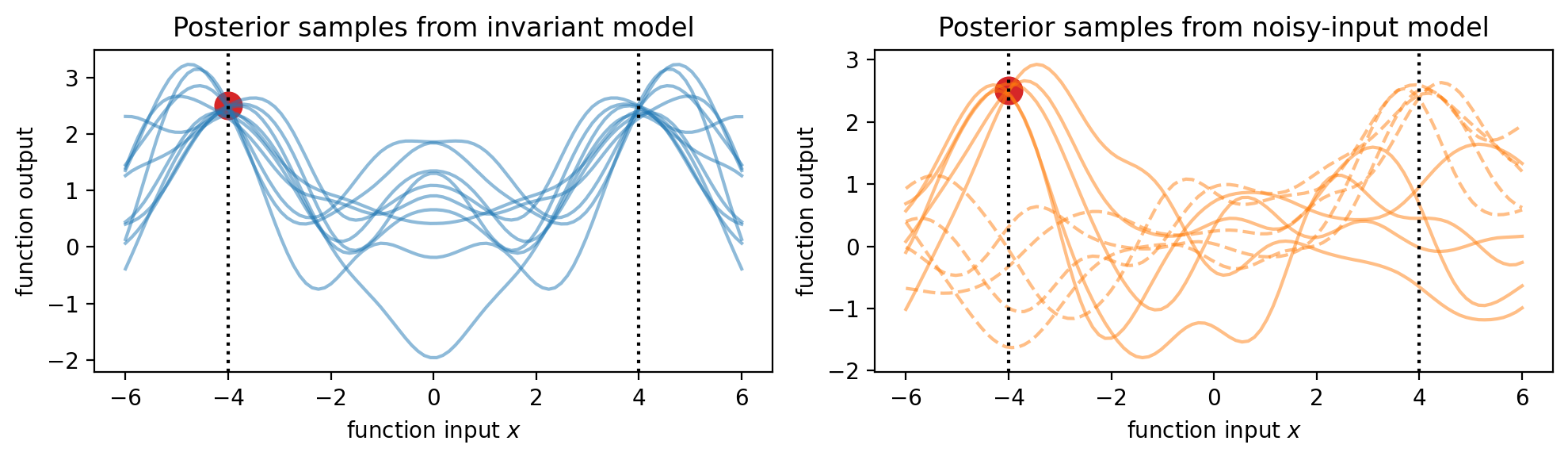}
    \caption{Samples from the model posteriors. The red dot marks the location of the observation $(x_1=-4,y_1=2.5)$. The noisy input posterior comprises two components: one conditioned on $(x_1,y_1)$ (dashed lines), the other on $(-x_1,y_1)$ (solid lines).}
    \label{fig:noisy_input_vs_invariance_samples}
\end{figure*}

In the noisy-input case, our observation is $(x_1,y_1)$, but $x$ is uncertain, so the observation could have been generated by $(-x_1,y_1)$ with equal probability. This results in a mixture posterior with two components: one component has ``seen'' $(x_1,y_1)$, while the other ``saw'' $(-x_1,y_1)$. The first component's prediction at $-x_1$ remains uninformed by its ``observation'' and the same is true for the second component's prediction at $x_1$. Thus, the predictions made by these components at these locations revert to the zero-mean prior. 

From the invariance perspective, we condition on the point $(x_1,y_1)$ but the double-sum kernel forces the function to be the same at $(-x_1,y_1)$. As the posterior is a single GP, it has unimodal marginals with high density around both points.

We can gain further intuition by looking at samples from both posteriors (Fig.~\ref{fig:noisy_input_vs_invariance_samples}). We can see that the \emph{every} sample from the invariance posterior is symmetric about $x=0$, where the functions drawn from the noisy input posterior are not symmetric in general.

The samples illustrate the key difference between the models. For the noisy input model, we can see the two components  of the mixture posterior arise from conditioning on different locations in the orbit of $x_1$ as described above. The component going through $(x=4,y_1)$ (samples drawn with dashed lines) is close to the prior at $(x=-4,y_1)$, the other (solid lines) goes through $(x=-4,y_1)$ and is close to the prior at $(x=4,y_1)$. However, under the invariance model, inference on the observation concentrates all model density around $y_1$ for both points in the orbit of $x_1$.

\begin{figure*}[t]
    \centering
    \begin{subfigure}{0.38\textwidth}
        \adjincludegraphics[trim={0 0 {.6\width} 0},clip,width=\textwidth]{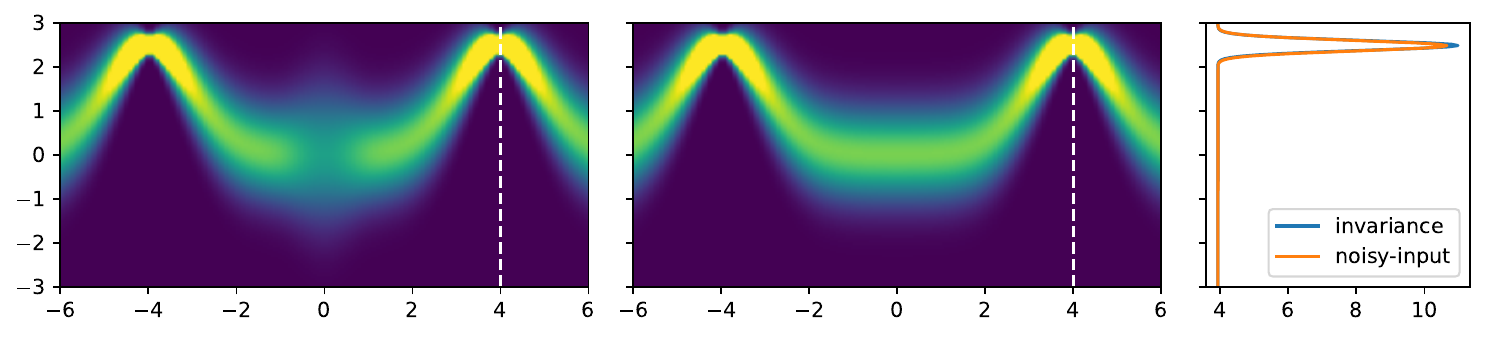}
        \subcaption{Invariant GP}
    \end{subfigure}%
    \begin{subfigure}{0.38\textwidth}
        \adjincludegraphics[trim={{.4\width} 0 {.2\width} 0},clip,width=\textwidth]{figs/app_general_lhoods/posterior_densities_10data.pdf}
       \subcaption{Noisy-input GP}
       \label{subfig:noisy_input_posterior_10data}
    \end{subfigure}%
    \begin{subfigure}{0.19\textwidth}
      \adjincludegraphics[trim={{.8\width} 0 0 0},clip,width=\textwidth]{figs/app_general_lhoods/posterior_densities_10data.pdf}
      \subcaption{$p(y|x=4,\mathcal{D})$}
      \label{subfig:noisy_input_vs_invariance_marginals_10data}
    \end{subfigure}%
    \caption{Posterior densities for the model constructions for ten observations at $\{(x_i=-4,y_i=2.5)\}_{i=1}^{10}$.}
    \label{fig:noisy_input_vs_invariance_densities_10data}
\end{figure*}

We now consider how this comparison changes as we observe more data. The noisy-input model \myeqref{eq:noisy_input_joint} requires integration over $\P(x'|x)$ to compute the likelihood of each datapoint, all of which are multiplied together to calculate their combined likelihood. Thus, the number of posterior components grows exponentially with the number of observations: $A^N$ (for orbit size $A$). Suppose all observations are at the same location $(x_1,y_1)$. In this case, the posterior density due to prior reversion at $\{x_1,-x_1\}$ decreases exponentially with $N$. This is because the fraction of mixture components conditioned on all input observations being at the same point in the orbit of $x_1$, i.e. all at $x_1$ or $-x_1$, is given by $A^{1-N}$. The predictive posteriors for ten observations, each at $(x=-4,y=2.5)$ is shown in Fig.~\ref{fig:noisy_input_vs_invariance_densities_10data}. Contrasting this noisy-input posterior (Fig.~\ref{subfig:noisy_input_posterior_10data}) to that for one observation (Fig.~\ref{subfig:noisy_input_posterior}), we can see the reduction in density around to the prior mean for points around the orbit of $x_1$. In summary, the noisy-input and invariance posteriors become more alike as we observe more data in the same orbit.

% TODO:
% \begin{itemize}
%     \item Example of different behaviours for exact orbit in GPs. Avg function gives strong generalisation to orbit, avg likelihood gives invariant predictive probability but with uncertainty along orbit.
%     \item How do marginal likelihoods compare for exact orbit GPs?
%     \item What does noisy input model posterior look like as training data grows?
%     \item Single-sample bound is equivalen to VI with posterior over input equal to the prior.
% \end{itemize}
\end{document}